\DocumentMetadata{}
\documentclass[sigconf,numbers,table]{acmart}
\usepackage{cleveref}
\usepackage{subcaption}
\usepackage[toc,page,title,titletoc]{appendix}

\usepackage{courier} 

\usepackage{multirow}
\usepackage{svg}
\usepackage{xcolor}
\usepackage{array}

\AtBeginDocument{%
  \providecommand\BibTeX{{%
    \normalfont B\kern-0.5em{\scshape i\kern-0.25em b}\kern-0.8em\TeX}}}

\copyrightyear{2023}
\acmYear{2023}
\setcopyright{licensedusgovmixed}\acmConference[SIGSPATIAL '24]{The 32nd ACM International Conference on Advances in Geographic Information Systems}{October 29--November 1, 2024}{Atlanta, GA, USA}
\acmBooktitle{The 32nd ACM International Conference on Advances in Geographic Information Systems (SIGSPATIAL '24), October 29--November 1, 2024, Atlanta, GA, USA}
\acmPrice{15.00}
\acmDOI{10.1145/3589132.3625590}
\acmISBN{979-8-4007-0168-9/23/11}

\begin{document}

\title{NUMOSIM: A Synthetic Mobility Dataset with Anomaly Detection Benchmarks}

\author{Chris Stanford}
\email{cstanford@novateur.ai}
\orcid{0000-0003-4437-7283}
\affiliation{%
  \institution{Novateur Research Solutions}
  \streetaddress{20110 Ashbrook Place Ste 170}
  \city{Ashburn}
  \state{Virginia}
  \country{USA}
  \postcode{20147}
}
\author{Suman Adari}
\email{sadari@novateur.ai}
\orcid{}
\affiliation{%
  \institution{Novateur Research Solutions}
  \streetaddress{20110 Ashbrook Place Ste 170}
  \city{Ashburn}
  \state{Virginia}
  \country{USA}
  \postcode{20147}
}
\author{Xishun Liao}
\email{xishunliao@ucla.edu}
\affiliation{%
  \institution{University of California, Los Angeles}
  \city{Los Angeles}
  \state{California}
  \country{USA}
  \postcode{90095}
}
\author{Yueshuai He}
\email{yueshuaihe@ucla.edu}
\affiliation{%
  \institution{University of California, Los Angeles}
  \city{Los Angeles}
  \state{California}
  \country{USA}
  \postcode{90095}
}
\author{Qinhua Jiang}
\email{qhjiang93@ucla.edu}
\affiliation{%
  \institution{University of California, Los Angeles}
  \city{Los Angeles}
  \state{California}
  \country{USA}
  \postcode{90095}
}
\author{Chenchen Kuai}
\email{kuai0407@ucla.edu}
\affiliation{%
  \institution{University of California, Los Angeles}
  \city{Los Angeles}
  \state{California}
  \country{USA}
  \postcode{90095}
}
\author{Jiaqi Ma}
\email{jiaqima@ucla.edu}
\affiliation{%
  \institution{University of California, Los Angeles}
  \city{Los Angeles}
  \state{California}
  \country{USA}
  \postcode{90095}
}

\author{Emmanuel Tung}
\email{etung@novateur.ai}
\orcid{}
\affiliation{%
  \institution{Novateur Research Solutions}
  \streetaddress{20110 Ashbrook Place Ste 170}
  \city{Ashburn}
  \state{Virginia}
  \country{USA}
  \postcode{20147}
}

\author{Yinlong Qian}
\email{yqian@novateur.ai}
\orcid{}
\affiliation{%
  \institution{Novateur Research Solutions}
  \streetaddress{20110 Ashbrook Place Ste 170}
  \city{Ashburn}
  \state{Virginia}
  \country{USA}
  \postcode{20147}
}

\author{Lingyi Zhao}
\email{lzhao@novateur.ai}
\orcid{}
\affiliation{%
  \institution{Novateur Research Solutions}
  \streetaddress{20110 Ashbrook Place Ste 170}
  \city{Ashburn}
  \state{Virginia}
  \country{USA}
  \postcode{20147}
}

\author{Zihao Zhou}
\email{ziz244@ucsd.edu}
\orcid{}
\affiliation{%
  \institution{University of California, San Diego}
  \city{San Diego}
  \state{California}
  \country{USA}
  \postcode{92093}
}

\author{Zeeshan Rasheed}
\email{zrasheed@novateur.ai}
\orcid{0000-0002-2369-9753}
\affiliation{%
  \institution{Novateur Research Solutions}
  \streetaddress{20110 Ashbrook Place Ste 170}
  \city{Ashburn}
  \state{Virginia}
  \country{USA}
  \postcode{20147}
}
\author{Khurram Shafique}
\email{kshafique@novateur.ai}
\orcid{0000-0002-3834-324X}
\affiliation{%
  \institution{Novateur Research Solutions}
  \streetaddress{20110 Ashbrook Place Ste 170}
  \city{Ashburn}
  \state{Virginia}
  \country{USA}
  \postcode{20147}
}

\renewcommand{\shortauthors}{Stanford, Adari, Liao, He, Jiang, Kuai, Ma, Tung, Qian, Zhao, Zhou, Rasheed, and Shafique}

\begin{abstract}


Collecting real-world mobility data is challenging. It is often fraught with privacy concerns, logistical difficulties, and inherent biases. Moreover, accurately annotating anomalies in large-scale data is nearly impossible, as it demands meticulous effort to distinguish subtle and complex patterns. These challenges significantly impede progress in geospatial anomaly detection research by restricting access to reliable data and complicating the rigorous evaluation, comparison, and benchmarking of methodologies. To address these limitations, we introduce a synthetic mobility dataset, NUMOSIM, that provides a controlled, ethical, and diverse environment for benchmarking anomaly detection techniques. NUMOSIM simulates a wide array of realistic mobility scenarios, encompassing both typical and anomalous behaviours, generated through advanced deep learning models trained on real mobility data. This approach allows NUMOSIM to accurately replicate the complexities of real-world movement patterns while strategically injecting anomalies to challenge and evaluate detection algorithms based on how effectively they capture the interplay between demographic, geospatial, and temporal factors. Our goal is to advance geospatial mobility analysis by offering a realistic benchmark for improving anomaly detection and mobility modeling techniques. To support this, we provide open access to the NUMOSIM dataset, along with comprehensive documentation, evaluation metrics, and benchmark results.

\end{abstract}



\maketitle
\vspace{-5pt}

\section{Introduction}

Mobility data analysis plays a critical role in various domains, including urban planning, disaster management, and epidemiology. Detecting anomalies within mobility patterns — deviations from expected behaviors — can provide valuable insights into significant events or disruptions. For instance, identifying mobility anomalies is essential for understanding the spread of infectious diseases and assessing the impact of public health interventions. However, research in geospatial anomaly detection is often constrained by the challenges associated with collecting and utilizing real-world mobility data. These challenges include stringent privacy regulations, high costs, logistical difficulties, and inherent biases in the data, all of which can impede the development and evaluation of robust anomaly detection algorithms.

Accurate annotation of anomalies in large-scale datasets presents another significant obstacle. Identifying subtle and complex patterns that indicate anomalies typically requires extensive domain expertise and meticulous effort. This process is not only time-consuming but also prone to inconsistencies, as human annotators may have different interpretations of what constitutes an anomaly. These limitations hinder access to reliable, well-annotated data, which is crucial for the rigorous evaluation and benchmarking of anomaly detection methodologies.

Given these constraints, synthetic data generation has emerged as a promising alternative for creating datasets that support the development and testing of anomaly detection algorithms. Synthetic datasets offer the flexibility to model a wide range of scenarios, including rare or extreme events that may be underrepresented in real-world data. However, generating synthetic mobility data that accurately captures the complexity of real-world movement patterns is challenging. A key difficulty lies in modeling the intricate interplay of demographic, geospatial, and temporal factors that shape human mobility behaviors. Traditional statistical approaches, typically employed in synthetic data generation, often fall short in capturing these complexities, resulting in datasets that lack the necessary variability and realism to effectively mimic real-world scenarios.

The remainder of this paper is organized as follows. Section 2 reviews prior work related to synthetic mobility data generation. Section 3 introduces the NUMOSIM data. Section 4 discusses the methodology used for generating NUMOSIM, including data modeling, model training, and the data generation processes. Section 5 details the anomaly injection processes, the types of anomalies introduced, and the rationale behind them. Section 6 describes the evaluation metrics and benchmark results provided with NUMOSIM. Section 7 presents our plans for expanding the dataset. Section 8 explains the data format and files released as part of the dataset. Finally, Section 9 concludes the paper with a discussion of contributions and future work.

\section{Prior Work}

There are numerous datasets available for studying human mobility, encompassing both real and synthetic data. These datasets can be categorized into three main types: location-based service (LBS) data, synthetic data, and vehicle data. 

\subsection{Location-Based Services (LBS) Data}
LBS data has been extensively used in recent studies of human mobility. For example, \citet{yabe2024yjmob100k} introduced YJMob100K, an open-source large-scale mobility dataset based on mobile phone data from Yahoo Japan Corporation. This dataset comprises 75 days of human mobility trajectories for 100,000 individuals, with location pings spatially and temporally discretized to protect user privacy. Other notable real-world datasets include Gowalla, which captures location-based social network data~\cite{Cho2011}, Foursquare, which includes check-ins from New York City and Tokyo from April 12, 2012, to February 13, 2013~\cite{yang_foursquare}, Meta’s travel pattern data, derived from the location history of Facebook users (Meta, n.d.), and OpenPaths from The New York Times Labs, which collects voluntary location data from iPhone and iPad users~\cite{openpaths_nytlabs}. However, LBS data often have limitations, such as incomplete daily trajectories, as seen with Gowalla and Foursquare, and a lack of detailed point of interest information due to privacy controls, as with YJMob100K. Additionally, these datasets typically do not include underlying sociodemographic characteristics of the agents because of privacy concerns. Another critical limitation is that real-world LBS datasets generally do not contain annotated anomalies, making it challenging to use them directly for developing and benchmarking anomaly detection algorithms.

\subsection{Synthetic Data}
The second category includes synthetic datasets, which offer an alternative by simulating human mobility patterns while overcoming some of the limitations of real-world data. For instance, HumoNet is a simulation framework that generates synthetic human trajectory data by leveraging multiple layers of real-world data, including transportation networks, points of interest (POI), population data, and observed human trajectories~\cite{humonet}. Another example is MetaPoL, a synthetic dataset created by simulating human behavior within an immersive virtual reality (VR) environment modeled after a real-world secure facility~\cite{metapol}. A significant advantage of synthetic datasets is their ability to provide fully annotated anomalies, allowing researchers to evaluate and benchmark anomaly detection algorithms effectively. This level of control and precision in data generation is difficult to achieve with real-world datasets. However, despite these benefits and innovative approaches, there is a notable lack of large-scale public synthetic datasets that can represent the complex dynamics of megacities with populations in the millions.

\subsection{Vehicle Data}
The final category is vehicle data, which primarily consists of datasets collected from GPS-equipped vehicles operating in urban environments. For instance, since 2009, the New York City Taxi \& Limousine Commission (NYC TLC) has released data on pick-up and drop-off locations for taxis and for-hire vehicles in New York City~\cite{nyc_tlc_data}. Other notable datasets include taxi datasets from cities such as Porto~\cite{moreira2013predicting}, San Francisco~\cite{epfl-mobility-20090224}, Rome~\cite{roma-taxi-20140717}, and Beijing~\cite{yuan2010t-drive}. The Porto dataset tracks 441 taxis over the course of one year~\cite{moreira2013predicting}, while the Rome dataset monitors 320 taxis over a 30-day period~\cite{roma-taxi-20140717, bonola2016opportunistic}. In addition to taxi datasets, there are also bus datasets that provide further insights into urban mobility. The Rio bus dataset~\cite{Bessa_Silva_Nogueira_Bertini_Freire_2016} and the Dublin bus dataset~\cite{Cruz_Barbosa_2023} capture the trajectories of buses across multiple staypoints or bus stops, allowing for the detection of abnormal mobility patterns. The anomalies are clearly defined as deviations from the established bus routes. Although these datasets offer valuable insights into the mobility patterns of taxi and bus users within their respective cities, they fall short of capturing complete daily trajectories and do not represent the broader urban population. This limitation restricts their applicability to more comprehensive mobility studies.

\section{Present Work}

To address the limitations of existing datasets, we introduce NUMOSIM (Novateur/UCLA Mobility Simulation), a novel synthetic mobility dataset designed to provide a controlled, ethical, and diverse environment for benchmarking anomaly detection techniques. NUMOSIM is generated using advanced generative deep learning models trained on real travel survey data (and optionally sparse mobility data, when available), enabling it to simulate a wide array of realistic mobility scenarios that encompass both typical and anomalous behaviors. Generative deep learning models are particularly well suited for this task, as they have demonstrated the ability to capture the complex interplay of demographic, geospatial, social, and temporal factors that drive human behavior, resulting in highly realistic synthetic data \cite{liao2024deep}. Our approach leverages these capabilities to simulate the trajectories of a large population of synthetic agents in urban areas. Additionally, NUMOSIM strategically injects a small quantity of generated anomalies into the dataset, facilitating rigorous benchmarking of anomaly detection algorithms by testing how effectively they capture the intricate dynamics of mobility patterns. 

To further support the development and refinement of these algorithms, we provide a comprehensive set of evaluation metrics and benchmark results alongside the NUMOSIM dataset. These resources are intended to help researchers assess the performance of their models and compare them with established baselines. 

The initial release of NUMOSIM focuses on a large-scale simulation of Los Angeles, resulting in NUMOSIM-LA — a dataset that reflects the unique mobility dynamics of this urban area. However, our vision for NUMOSIM extends beyond a single city. We plan to expand the dataset by continuing to add more cities in the future, beginning with NUMOSIM-CAIRO, to increase its utility and applicability across different geographic contexts. We aim to advance geospatial mobility analysis by offering a realistic and versatile benchmark for testing and improving anomaly detection and mobility modeling techniques.

\section{Methodology}

The Deep Activity Model (DeepAM)~\cite{liao2024deep} introduces a novel approach to the synthesis of human mobility patterns by harnessing the capabilities of deep learning techniques and the widespread availability of household travel survey (HTS) data. This model employs a transformer-based architecture, featuring an encoder-decoder structure, to generate realistic activity chains that capture the complex interdependencies among household members and their activities.

The input of the model consists of sociodemographic data of individuals and their household members. These diverse data types are concatenated and embedded to form a comprehensive feature vector. The encoder processes this input through self-attention layers, effectively capturing the intricate relationships within the data. Subsequently, the decoder generates activity predictions in an autoregressive manner, predicting the activity type, start time, and end time for each activity in an activity chain.

DeepAM is trained on 180,000 samples from the 2017 National Household Travel Survey (NHTS)~\cite{nhtsdata}. The activities used in the NHTS informed our activity classification into the 16 categories shown in Table~\ref{table:activitytypes}. The model's loss function incorporates multiple components: cross-entropy loss for activity-type prediction, soft-label loss for temporal predictions, and specialized losses for temporal order and sequential timing to ensure logically consistent chronological sequences.

\begin{table}[h!]
\centering
\begin{tabular}{>{\raggedright\arraybackslash}p{3.5cm} >{\raggedright\arraybackslash}p{3.5cm}}
    \rowcolor{gray!40} \multicolumn{2}{c}{\textbf{Activity Types}} \\ 
    \rowcolor{white} Transportation & Errands \\
    \rowcolor{gray!10} Home & Recreation \\
    \rowcolor{white} Work & Exercise \\
    \rowcolor{gray!10} School & Visit \\
    \rowcolor{white} ChildCare & HealthCare \\
    \rowcolor{gray!10} BuyGoods & Religious \\
    \rowcolor{white} Services & SomethingElse \\
    \rowcolor{gray!10} EatOut & DropOff \\
\end{tabular}
\caption{List of the 16 modeled activity types.}
\label{table:activitytypes}
\end{table}

Given the inherent uncertainty of human behavior, the performance and realism of DeepAM are assessed at a systemic level by comparing the distributions of generated and real-world activity patterns. Specifically, Jensen-Shannon divergence (JSD) is used to evaluate five key aspects of the generated activity chains: activity frequencies, distributions of activity start and end times, the number of daily activities, and activity durations. In addition, the logical consistency of the model is evaluated through the activity-transition probability similarity. The results of these evaluations can be found in~\cite{liao2024deep}.

For this release, activity type chains were generated for a synthetic subpopulation of Los Angeles consisting of 200,000 agents. An example of an activity chain for a single agent is shown in Table~\ref{table:activitydata}.

\begin{table}[h!]
\centering
\begin{tabular}{>{\raggedright\arraybackslash}p{1cm} >{\raggedright\arraybackslash}p{2cm} >{\raggedright\arraybackslash}p{2cm} >{\raggedright\arraybackslash}p{2cm}}
    \rowcolor{gray!40} \textbf{Day} & \textbf{Activity} & \textbf{Start time} & \textbf{End time} \\
    \rowcolor{white} 1 & Home & 00:00 & 08:30 \\
    \rowcolor{gray!10} 1 & Work & 09:00 & 17:00 \\
    \rowcolor{white} 1 & Home & 17:30 & 24:00 \\
    \rowcolor{gray!10} 2 & Home & 00:00 & 09:00 \\
    \rowcolor{white} 2 & Work & 09:30 & 12:00 \\
    \rowcolor{gray!10} 2 & EatOut & 12:15 & 12:45 \\
    \rowcolor{white} 2 & Work & 13:00 & 17:30 \\
    \rowcolor{gray!10} 2 & Home & 18:00 & 24:00 \\
    \rowcolor{white} \dots & \dots & \dots & \dots \\
    \rowcolor{gray!10} 56 & Home & 17:45 & 24:00 \\
\end{tabular}
\vspace{5pt}
\caption{Example of a single agent's 8-week activity chain output from the Deep Activity Model.}
\label{table:activitydata}
\end{table}

\begin{figure}[h]
  \centering
  \includegraphics[width=0.48\textwidth]{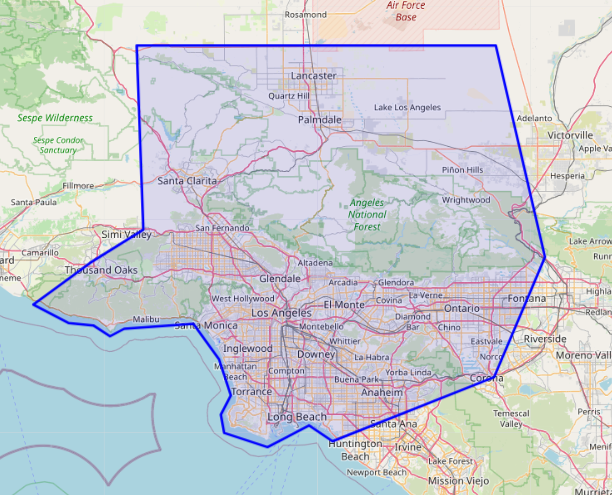}
  \caption{Area of interest covered in the simulation. All agents stay within the boundary during the 8 week period. Map data from OpenStreetMap~\cite{openstreetmap}.}
  \label{fig:aoi}
\end{figure}

The list of available Points of Interest (POIs) for the simulation was curated using data from Planetsense\cite{planet_sense} and USA Structures\cite{fema_structures} for the Los Angeles region, as defined by the boundary in \Cref{fig:aoi}. POIs were matched to nearby structures, with unmatched structures assumed to be residential locations. Each location was then assigned a set of valid activity types based on a manually defined mapping. For example, locations labeled "education" by Planetsense were assigned activity types such as Work, School, and DropOff. Residential locations were assigned activity types like Home and Visit. The total number of POIs available for each activity type is presented in \Cref{table:poicounts}.

\begin{table}[h!]
\centering
\begin{tabular}{>{\raggedright\arraybackslash}p{3cm} >{\raggedright\arraybackslash}p{3cm}}
    \rowcolor{gray!40} \textbf{Activity Type} & \textbf{Number of POIs} \\
    \rowcolor{white} Transportation & 449 \\
    \rowcolor{gray!10} Home & 2509756 \\
    \rowcolor{white} Work & 409920 \\
    \rowcolor{gray!10} School & 10904 \\
    \rowcolor{white} ChildCare & 33821 \\
    \rowcolor{gray!10} BuyGoods & 108496 \\
    \rowcolor{white} Services & 17028 \\
    \rowcolor{gray!10} EatOut & 165442 \\
    \rowcolor{white} Errands & 4414 \\
    \rowcolor{gray!10} Recreation & 17685 \\
    \rowcolor{white} Exercise & 30520 \\
    \rowcolor{gray!10} Visit & 2509756 \\
    \rowcolor{white} HealthCare & 3100 \\
    \rowcolor{gray!10} Religious & 7255 \\
    \rowcolor{white} SomethingElse & 2054 \\
    \rowcolor{gray!10} DropOff & 2838192 \\
\end{tabular}
\vspace{5pt}
\caption{Number of POIs in the area of interest that correspond to each activity type. A given POI can correspond to multiple activity types.}
\label{table:poicounts}
\end{table}

Following the generation of activity type chains by DeepAM, a POI assignment procedure was employed to assign each activity to a valid POI for that activity. This procedure considered available data and literature on LA, including average commute times ~\cite{yardikube_commuting}, total distance traveled per day~\cite{Noulas2012, Alessandretti2017}, radius of gyration~\cite{Gonzalez2008, Pappalardo2015}, number of locations visited per day~\cite{Schneider2013}, and Zipf's law human movement~\cite{Gonzalez2008,Zipf1946}. Once agents were assigned POIs, they navigated the Los Angeles road network, obtained via Open Street Map (OSM)~\cite{openstreetmap} to perform their activities. Their arrival times are results of their navigation through the road network using the speed limits supplied by OSM. Although we did not include second-by-second trajectories in the data release, this step is necessary to generate realistic arrival and departure times.

The final output is a schedule of POI visits for each agent, spanning two consecutive 4-week periods: \texttt{train} and \texttt{test}. An example of the format for the train POI visits is shown in Table~\ref{table:poidata}.

\begin{table}[t!]
\centering
\begin{tabular}{>{\raggedright\arraybackslash}p{1cm} >{\raggedright\arraybackslash}p{1cm} >{\raggedright\arraybackslash}p{2.5cm} >{\raggedright\arraybackslash}p{2.5cm}}
    \rowcolor{gray!40} \textbf{agent} & \textbf{poi\_id} & \textbf{start\_datetime} & \textbf{end\_datetime} \\
    \rowcolor{white} 1 & 215521 & 2024-01-01 T00:00:00-08:00 & 2024-01-01 T08:29:09-08:00 \\
    \rowcolor{gray!10} 1 & 89362 & 2024-01-01 T09:01:01-08:00 & 2024-01-01 T17:05:11-08:00 \\
    \rowcolor{white} 1 & 215521 & 2024-01-02 T08:35:10-08:00 & 2024-01-0 2T09:00:55-08:00 \\
    \rowcolor{gray!10} 1 & 89362 & 2024-01-02 T09:31:05-08:00 & 2024-01-02 T12:00:05-08:00 \\
    \rowcolor{white} 1 & 99721 & 2024-01-02 T12:12:44-08:00 & 2024-01-02 T12:41:31-08:00 \\
    \rowcolor{gray!10} 1 & 89362 & 2024-01-02 T13:02:00-08:00 & 2024-01-02 T17:36:52-08:00 \\
    \rowcolor{white} 1 & 215521 & 2024-01-02 T18:03:20-08:00 & 2024-01-03 T08:16:30-08:00 \\
    \rowcolor{gray!10} \dots & \dots & \dots & \dots \\
    \rowcolor{white} 200000 & 788111 & 2024-01-28 T17:34:50-08:00 & 2024-01-29 T08:16:21-08:00 \\
\end{tabular}
\caption{Example of the stay\_points\_train.parquet data format, containing the stay points for 200,000 agents for 4 weeks.}
\label{table:poidata}
\end{table}

\begin{table*}[t!]
\centering
\begin{tabular}{>{\raggedright\arraybackslash}p{3.5cm} >{\raggedright\arraybackslash}p{2cm} >{\raggedright\arraybackslash}p{2cm} | >{\raggedright\arraybackslash}p{3.5cm} >{\raggedright\arraybackslash}p{2cm} >{\raggedright\arraybackslash}p{2cm}}
    \rowcolor{gray!40} 
    \multicolumn{3}{c}{\textbf{Original}} & \multicolumn{3}{c}{\textbf{Anomalous}} \\
    \rowcolor{gray!40} 
    \textbf{poi\_id} & \textbf{start\_datetime} & \textbf{end\_datetime} & \textbf{poi\_id} & \textbf{start\_datetime} & \textbf{end\_datetime} \\
    
    \rowcolor{white} 2364463\newline (residence) & 2024-02-07\newline16:38:04 & 2024-02-08\newline07:17:01 &
    2364463\newline (residence) & 2024-02-07\newline16:38:04 & \textcolor{red}{2024-02-08\newline01:48:01} \\
    
    \rowcolor{gray!10} -& & -&
    \textcolor{red}{295533\newline Silver Screen Sight and Sound} & \textcolor{red}{2024-02-08\newline02:28:56} & \textcolor{red}{2024-02-08\newline13:19:47} \\
    
    \rowcolor{white} 399191\newline Ramona Elementary School & 2024-02-08\newline07:33:18 & 2024-02-08\newline17:41:43 &
    399191\newline Ramona Elementary School & \textcolor{red}{2024-02-08\newline14:01:34} & 2024-02-08\newline17:41:43 \\

    \rowcolor{white} 2364463\newline (residence) & 2024-02-08\newline17:58:47 & 2024-02-09\newline06:55:00 &
    2364463\newline (residence) & 2024-02-08\newline17:58:47 & 2024-02-09\newline06:55:00 \\
\end{tabular}
\caption{Tabular example of an injected anomaly, corresponding to \Cref{fig:map_normal_vs_anomalous} and \Cref{fig:recurring} (Test Week 2, Thursday). Anomalies (red) are created as a result of the injected visit, as well as the temporal adjustments to surrounding visits.}
\label{table:anomalous}
\end{table*}

\begin{figure*}[h]
  \centering
  
  \vspace{20pt} 
  \hspace*{-6mm}
  \includegraphics[width=1.04\textwidth]{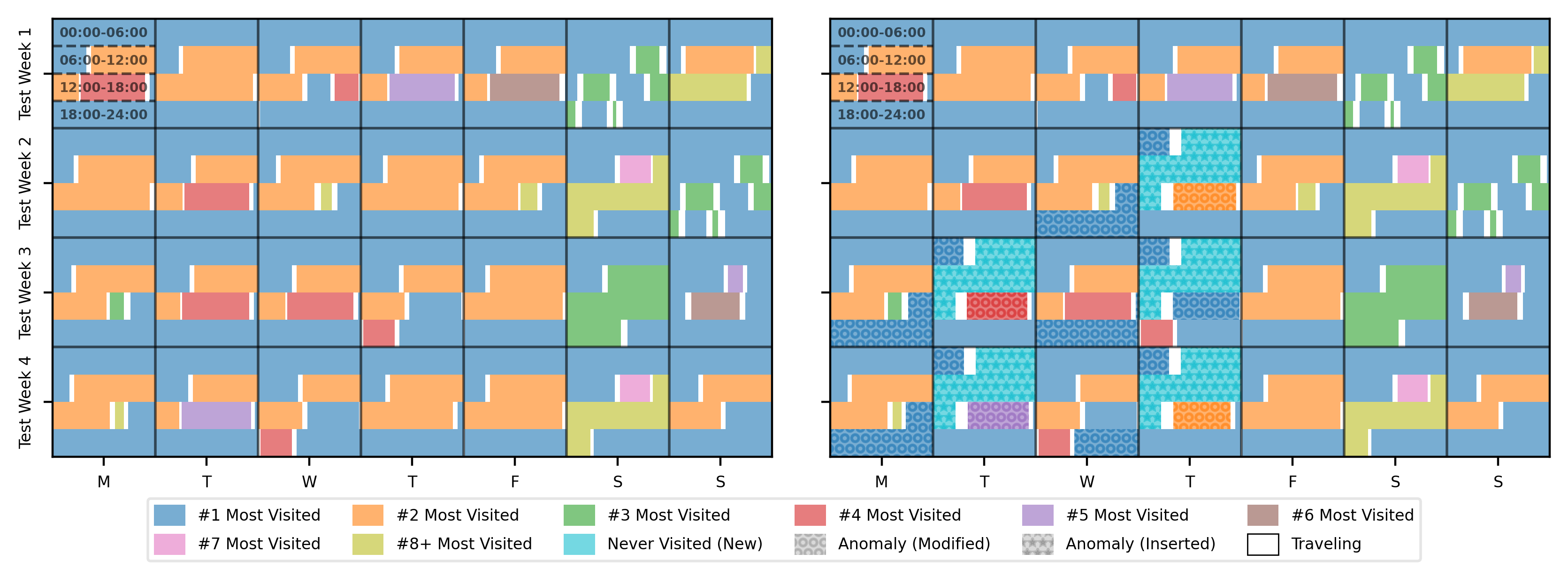}
  \caption{Example of a recurring anomaly. Left: the unaltered, normal test period for an agent, represented as a calendar. Each box represents one day, and each row within a box represents a six-hour period (as demonstrated in the top-left box). Right: the altered, anomalous test period for the same agentThe injected anomalies (marked with the star hatch pattern) are a visit to a new location (cyan) for that agent, recurring at the same time of day each time. The surrounding visits are also altered temporally to accommodate and are therefore also considered anomalous.}
  \label{fig:recurring}
\end{figure*}

\begin{figure*}[!ht]
    \vspace{20pt} 
    \centering
    \begin{subfigure}{0.485\textwidth}
        \centering
        \includegraphics[width=\linewidth]{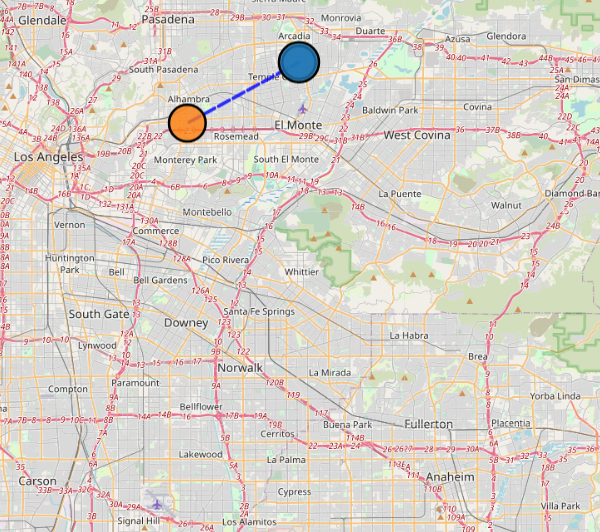}
        \caption{Normal daily activity sequence. The agent starts at home (blue), then goes to a frequently visited location (orange), and then returns home.}
    \end{subfigure}
    \hfill
    \vspace{5pt} 
    \begin{subfigure}{0.485\textwidth}
        \centering
        \includegraphics[width=\linewidth]{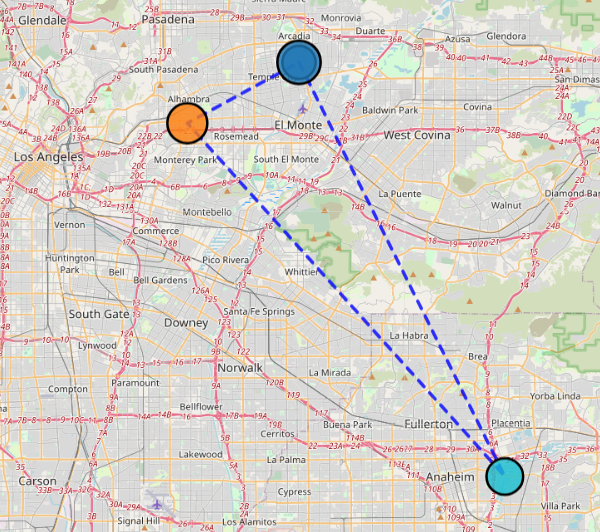}
        \caption{Anomalous daily activity sequence. A visit at a new location (cyan) is inserted, making that visit anomalous, as well as the surrounding points which are altered temporally to accommodate.}
    \end{subfigure}
    \caption{Example of normal (Left) and anomalous (Right) daily stay point sequence for the same agent as \Cref{fig:recurring} on the day of Test Week 2, Day 4 (Thursday).}
    \label{fig:map_normal_vs_anomalous}
\end{figure*}

\section{Anomaly Injection}


The NUMOSIM dataset includes two primary types of anomalies: non-recurring anomalies and recurring anomalies. These anomalies are strategically injected into the dataset to simulate different deviations from expected agent behavior patterns, providing a robust benchmark for evaluating the effectiveness of anomaly detection algorithms in geospatial mobility data.

Non-recurring anomalies are singular, isolated events that disrupt an agent's typical sequence of activities during the test period. These anomalies are designed to represent unexpected, one-time deviations from an agent's routine. For example, an agent might unexpectedly leave a stay point that they typically visit at a specific time, or they might visit a new location that fundamentally differs from their usual pattern of stay points. These anomalies challenge detection models to identify significant deviations without being misled by the inherent variability of human behavior.

Recurring anomalies, on the other hand, represent patterns of behavior that deviate from the norm on a regular basis. These anomalies are injected to simulate shifts in an agent's established behavioral patterns, such as changes in the frequency, timing, or location of visits to Points of Interest (POIs). For example, an agent who previously visited a particular location once a month might now visit it every weekend. While each visit might seem normal in isolation, the recurring nature of these visits marks a significant departure from the agent's typical behavior, thereby classifying it as an anomaly. Another instance of a recurring anomaly could involve an agent routinely visiting a new location instead of their usual spot. 

\Cref{fig:recurring}, \Cref{fig:map_normal_vs_anomalous}, and \Cref{table:anomalous} for illustrate an example of a recurring anomaly. \Cref{fig:recurring} shows a calendar-based visualization of a normal 4-week period compared to an anomalous 4-week period. \Cref{fig:map_normal_vs_anomalous} shows a map-based visualization of the spatial activity of a particular day from \Cref{fig:recurring}, which is further specified by \Cref{table:anomalous}.

The primary objective of introducing these anomalies is to assess how well detection models can differentiate between normal behavioral variations and true anomalies. An effective model should be capable of recognizing both subtle and overt deviations from established patterns, minimizing false positives by correctly distinguishing between normal but irregular behaviors and genuinely anomalous events. This capability is crucial for applications where the ability to identify real disruptions - such as those caused by emergencies, unexpected events, or behavioral changes - is essential.  Non-recurring anomalies test a model's sensitivity to singular disruptions, ensuring that the model does not overlook important one-time deviations. In contrast, recurring anomalies evaluate a model's ability to recognize shifts in behavior that may be less obvious but are significant due to their repetitive nature. Together, these anomalies provide a comprehensive assessment of model robustness and accuracy, making NUMOSIM a powerful tool for benchmarking anomaly detection in geospatial mobility data.




\section{Benchmarks}

To demonstrate the utility of NUMOSIM as a comprehensive testbed for geospatial anomaly detection, we implement a set of baseline methods for benchmarking. These benchmarks are designed not only to validate the detectability of the injected anomalies but also to establish a performance baseline for future work in this domain.

NUMOSIM, with its labeled anomalous and non-anomalous data, provides a versatile platform for developing and evaluating anomaly detection models. The dataset's flexibility allows for various configurations, including both unsupervised and semi-supervised approaches. The anomaly prevalence rate during evaluation can be adjusted by selecting specific subsets of the \texttt{test} sers can introduce various types of noise, such as Gaussian spatial/temporal noise, missing staypoints, or agent ID switches to tailor the data to specific application needs.

Beyond data partitioning and augmentation, NUMOSIM allows for training and evaluation at different levels of granularity: agent-level, staypoint-level, trip-level, or point-level. For instance, an agent may be classified as anomalous if it has a certain number or proportion of anomalous staypoints; this classification can then be used to score models at the agent-level. Similarly, models that output scores for staypoints or trips (i.e., staypoint-pair transitions) can be evaluated at those levels, or aggregated to the agent-level using strategies such as max-pooling staypoint scores. For applications sensitive to duration, evaluations can be conducted on a second-by-second or point-by-point basis, allowing for fine-grained analysis of model performance.

Our benchmarking methodology proceeds as follows. First, we train the selected methods on the \texttt{train} dataset, which is free of anomalies, using an unsupervised learning approach. Next, we perform inference on the \texttt{test} period data to generate anomaly scores. Finally, we evaluate each model using metrics such as the Average Precision (AP) and Area Under the Receiver Operating Characteristic curve (AUCROC). Although other metrics like Maximum F1-Score or Average Precision Recall are viable, they not are included in this analysis. It is important to note that no forms of noise or data augmentation were applied during this benchmarking process.

We selected three baseline anomaly detection methods for this evaluation: RioBusData \citep{Bessa_Silva_Nogueira_Bertini_Freire_2016}, Spatial-Temporal Outlier Detector (STOD) \citep{Cruz_Barbosa_2020}, and Gaussian Mixture Variational Sequence AutoEncoder (GM-VSAE) \citep{Liu_Zhao_Cong_Bao_2020}. RioBusData \citep{Bessa_Silva_Nogueira_Bertini_Freire_2016} is an interactive analysis tool with a CNN-based anomaly detection model, initially applied to bus mobility data. STOD \citep{Cruz_Barbosa_2020} is a GRU-based anomaly detection model, also originally designed for bus mobility data. GM-VSAE \citep{Liu_Zhao_Cong_Bao_2020} is a VAE-based anomaly detection model, applied to taxi mobility data. To adapt these baseline methods to NUMOSIM, we made a few straightforward modifications, which are detailed in Appendix~\ref{appendix:baseline_method_adaptations}. In addition to the baseline methods, we created a simple statistical test, described in Appendix~\ref{appendix:visit_rate_model}, based on the fact that the rate at which agents visit each POI should not change substantially between train and test. 

The results of our baseline benchmarking experiments are presented in \Cref{table:benchmarks}. In this table, we observe a striking trend: most existing, sophisticated anomaly detection models are underperforming, while a basic model that merely accounts for the visit rate of agents achieves significantly higher scores. 

This outcome, while surprising, is not unprecedented in the field of anomaly detection. In \citet{Wu2020}, they demonstrate that with just a single line of code, they can replicate the performance of complex, many-parameter deep learning models on popular benchmark datasets.

This observation underscores the broader principle that the success of anomaly detection, particularly in the context of complex mobility patterns, hinges not just on the sophistication of the models employed but on the careful selection of features that truly capture the underlying dynamics of the data. In mobility analysis, where patterns can be influenced by a myriad of factors such as time, location, sociodemographic characteristics, and behavioral trends, identifying the most relevant features is crucial. A well-chosen feature, like the visit rate, can often encapsulate significant behavioral signals, providing clarity where more complex models might struggle to disentangle noise from meaningful patterns.

\begin{table}[t!]
\centering
\begin{tabular}{ll>{\raggedright\arraybackslash}p{1.6cm}>{\raggedleft\arraybackslash}p{1.1cm}>{\raggedleft\arraybackslash}p{1.2cm}}
    \rowcolor{gray!40} \textbf{Level} & \textbf{Method} & \textbf{Anomaly Prevalence} & \textbf{AP} & \textbf{AUCROC} \\
    \multirow{4}{*}{Staypoint} & RioBusData & \multirow{4}{*}{0.000203} & 0.000189 & 0.505 \\
    & STOD &  & 0.00024 & 0.594 \\
    & GM-VSAE &  & --- & --- \\
    & \textbf{Visit Rate} &  & \textbf{0.0165} & \textbf{0.906} \\
    \cellcolor{gray!10} & \cellcolor{gray!10} RioBusData & \cellcolor{gray!10} & \cellcolor{gray!10}0.00164 & \cellcolor{gray!10}0.501 \\
    \cellcolor{gray!10} & \cellcolor{gray!10} STOD & \cellcolor{gray!10} & \cellcolor{gray!10}0.00182 & \cellcolor{gray!10}0.518 \\
    \cellcolor{gray!10} & \cellcolor{gray!10} GM-VSAE & \cellcolor{gray!10} & \cellcolor{gray!10}0.00192 & \cellcolor{gray!10}0.507 \\
    \multirow{-4}{*}{\cellcolor{gray!10}Agent} & \cellcolor{gray!10} \textbf{Visit Rate} & \multirow{-4}{*}{\cellcolor{gray!10}0.00191} & \cellcolor{gray!10}\textbf{0.0164} & \cellcolor{gray!10}\textbf{0.646} \\
\end{tabular}
\caption{Staypoint-level and agent-level results from baseline methods on NUMOSIM-LA. GM-VSAE concatenates an agent's entire series of stay points, so there are no staypoint-level scores for this method.}
\label{table:benchmarks}
\end{table}

\section{Ongoing Releases}

In addition to the initial Los Angeles dataset, we have plans for several future releases that will be made available as they are completed. Our next planned release will focus on Cairo, Egypt, with additional regions to follow over time. Each new dataset will be tailored to reflect the unique mobility patterns and cultural contexts of its respective region.

To achieve this level of specificity, we will employ transfer learning techniques, enabling our deep activity model to adapt to the local customs and behaviors of each area. By incorporating local human mobility data into the training process, our model will be fine-tuned to capture the distinct activity patterns that characterize different regions, ensuring that each dataset offers a realistic and region-specific representation of human movement.

These ongoing releases are part of our broader commitment to providing comprehensive, high-quality datasets that can be used to benchmark and refine geospatial mobility models and anomaly detection techniques. By expanding the geographic scope of our datasets, we not only aim to support a wider range of research applications—from urban planning and transportation management to public health and security—but also to enhance the generalizability of anomaly detection models across different geospatial regions \cite{tenzer_generalization2023}. This approach will allow researchers to test and validate their models in diverse environments, ensuring that the developed techniques are robust and applicable across various contexts. By offering datasets that span multiple regions, we hope to contribute to the creation of more versatile and adaptable models capable of handling the complexities of human mobility on a global scale.

\section{File Descriptions}

NUMOSIM data can be accessed on the Open Science Framework at (\href{https://osf.io/sjyfr/}{https://osf.io/sjyfr/}).

This release contains 6 files, split into two categories: supplemental and stay point.

\subsection{Supplemental Files}

\begin{enumerate}
    \item \textbf{readme.txt}: A text file describing the files in the data release.
    \item \textbf{demographics.parquet}: Table containing the demographic information for each agent. Descriptions of the columns and their values can be found in the readme file.
    \item \textbf{poi.parquet}: Table containing information for points of interest. Columns are poi\_id, name, latitude, and longitude, and act\_types. 
\end{enumerate}
\subsection{Stay point files}

The stay point files contain the stay point information for the agents. The 4-week \texttt{train} stay points and 4-week \texttt{test} stay points are split into separate files, but together comprise an 8-week sequence of stay points for 200,000 simulated agents, spanning a period from 2024-01-01T00:00:00-08:00 to 2024-02-25T23:59:59-08:00. Any activity that overlapped the beginning of \texttt{train} or the end of \texttt{test} was truncated. Any activity that overlapped the transition between \texttt{train} and \texttt{test} was included in the \texttt{train} file in full rather than being split across the two files. 

\begin{enumerate}
    \item \textbf{stay\_points\_train.parquet}: Table containing stay point information for all agents for a 4-week long training period. Columns are agent\_id, poi\_id, start\_datetime, and end\_datetime.
    \item \textbf{stay\_points\_test\_truth.parquet}: Table containing ground truth stay point information for all agents for a 4-week long period that immediately follows the training period. Columns are agent\_id, poi\_id, start\_datetime, and end\_datetime.
    \item \textbf{stay\_points\_test\_anomalous.parquet}: Table containing stay point information for all agents for a 4-week long period that immediately follows the training period, with injected anomalies. Columns are agent\_id, poi\_id, start\_datetime, end\_datetime, anomaly, and anomaly\_type. The anomaly column is a boolean indicating whether the stay point has been injected or altered. The anomaly\_type column is a integer where 1 indicates that the anomaly is a modified staypoint and 2 indicates that the anomaly is an injected staypoint.
\end{enumerate}


\section{Conclusion}

In this paper, we introduced NUMOSIM, a synthetic mobility dataset with embedded anomalies, designed as a benchmark for evaluating geospatial mobility models and anomaly detection techniques. By leveraging deep activity models trained on real-world survey data, we aim to bridge the gap between synthetic data generation and the realistic representation of human mobility patterns.

This dataset represents a significant step forward in the field of geospatial mobility analysis, providing researchers with a robust tool to develop and validate more sophisticated models that can generalize beyond the limitations of traditional datasets. Our future work will focus on expanding the dataset to encompass different geographical regions, incorporating more complex and rare behavioral patterns and social interactions, and refining the anomaly generation process to further enhance the realism and utility of the dataset.

By making this dataset publically available, we aim to drive progress in mobility modeling and anomaly detection, ultimately contributing to more accurate and effective applications in areas such as urban planning and transportation management.

\vspace{-5pt}
\begin{acks}
Supported by the Intelligence Advanced Research Projects Activity (IARPA) via Department of Interior/Interior Business Center (DOI/IBC) contract number 140D0423C0033. The U.S. Government is authorized to reproduce and distribute reprints for Governmental purposes notwithstanding any copyright annotation thereon. Disclaimer: The views and conclusions contained herein are those of the authors and should not be interpreted as necessarily representing the official policies or endorsements, either expressed or implied, of IARPA, DOI/IBC, or the U.S. Government. 

\end{acks}
\vspace{-5pt}
\bibliographystyle{ACM-Reference-Format}
\bibliography{paper}

\begin{appendices}
\section{Baseline Method Adaptations}
\label{appendix:baseline_method_adaptations}

In this section, we describe how we modify the Rio-Bus Model~\cite{Bessa_Silva_Nogueira_Bertini_Freire_2016}, the STOD~\cite{Cruz_Barbosa_2020} and the GM-VSAE~\cite{Liu_Zhao_Cong_Bao_2020} to use with our multi-agent staypoint trajectories dataset.

\begin{figure}[htbp]
    \centering
    \includegraphics[width=0.8\columnwidth]{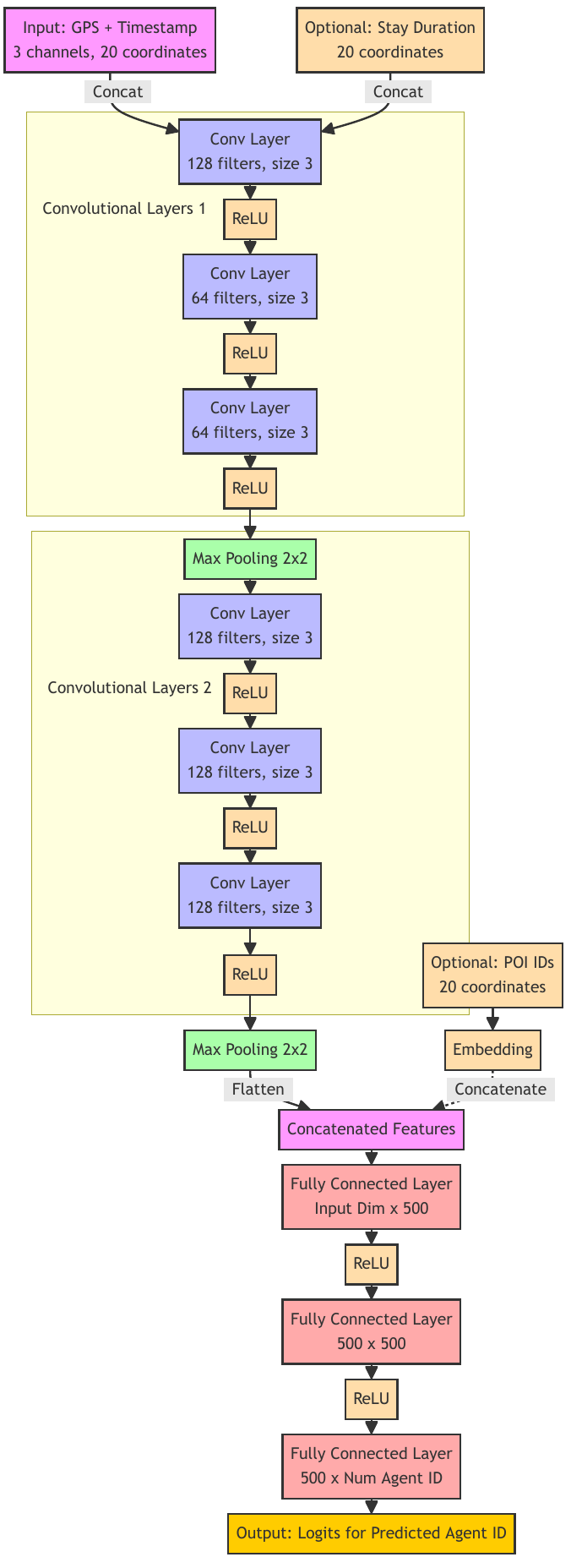}
    \caption{Illustration of the RioBUS CNN architecture adapted to the multi-agent GPS staypoint trajectory. The input to the network is 20 GPS coordinates of a single agent that are coded in three channels, which correspond to the timestamp, latitude, and longitude of each GPS coordinate. The input can be optionally extended to four channels by the additional stay duration feature. This input is fed to two convolutional layer blocks. The CNN output is then concatenated with an optional POI embedding. The last layer predicts the agent ID to which the input 20 GPS coordinates belong.}
    \label{fig:riobus-flow}
\end{figure}

\subsection{RioBusData}

The original RioBusData model~\cite{Bessa_Silva_Nogueira_Bertini_Freire_2016} uses a Convolutional Neural Network (CNN) to analyze bus trajectories in Rio de Janeiro. The CNN takes as input a sequence of GPS coordinates and predicts the bus ID, learning to distinguish between normal and outlier trajectories by assuming outlier bus trajectories' IDs are harder to predict. The architecture of the adapted model is presented in Figure \ref{fig:riobus-flow}.

\paragraph{Agent ID Replacement}
The bus ID in the original model is replaced by an agent ID. This allows the model to build agent behaviors' classification instead of bus routes' classification.

\paragraph{Additional Features}
Points of Interest (POI) embeddings are concatenated with the CNN output before the final prediction layer. This provides context about the surrounding environment of each GPS coordinate. Additionally, a new channel is added to the input, representing the stay duration at each GPS coordinate. This helps capture the temporal aspect of staypoints.

\paragraph{Input Structure}
The adapted model still takes as input 20 GPS coordinates for a single agent. These are encoded in three or four channels: timestamp, latitude, longitude, and optionally, stay duration.

\paragraph{Network Architecture}
The core CNN structure remains similar to the original RioBusData model. It consists of three initial convolutional layers, followed by a max pooling layer, then three more convolutional layers, another max pooling layer, and finally three fully connected layers.

\paragraph{Output Score}
We use the negative probability of the agent ID being correctly predicted as the anomaly score. If the agent's sub-trajectory is less likely to be correctly identified, it is more likely to be anomalous. The staypoint-level anomaly score is computed as the anomaly score of the 20-coordinates sub-trajectory ended with the staypoint. The agent-level anomaly score is computed from all 20-coordinate sub-trajectories of a specific agent. It is equal to the maximum score among these sub-trajectories.

\subsection{STOD}

The original STOD model~\cite{Bessa_Silva_Nogueira_Bertini_Freire_2016} applies a three-stage approach. The model first employs a bidirectional GRU (Gated Recurrent Unit) to process contextual information. This includes historical trajectory, future trajectory, and other features of the coordinate of interest. The GRU then predicts the bus state at the coordinate of interest. The bus state is one of ``in route, bus stop, trafﬁc signal, and other stops''. The classifier is called PAC (Point Activity Classiﬁer). The high dimensional input to the last layer of the PAC is considered a contextual representation of a point in the trajectory.

Meanwhile, it maps all GPS sequences to H3~\cite{H3_2018} hexagonal grids. All GPS coordinates are converted to discrete words representing the grid index. A Word2Vec~\cite{Church_2017} model is trained on the H3 sequences to gather the association between H3 grids. The Word2Vec embeddings of a GPS coordinate are considered a geographical representation of a point.

\begin{figure*}[htbp]
    \centering
    \includegraphics[width=0.9\textwidth]{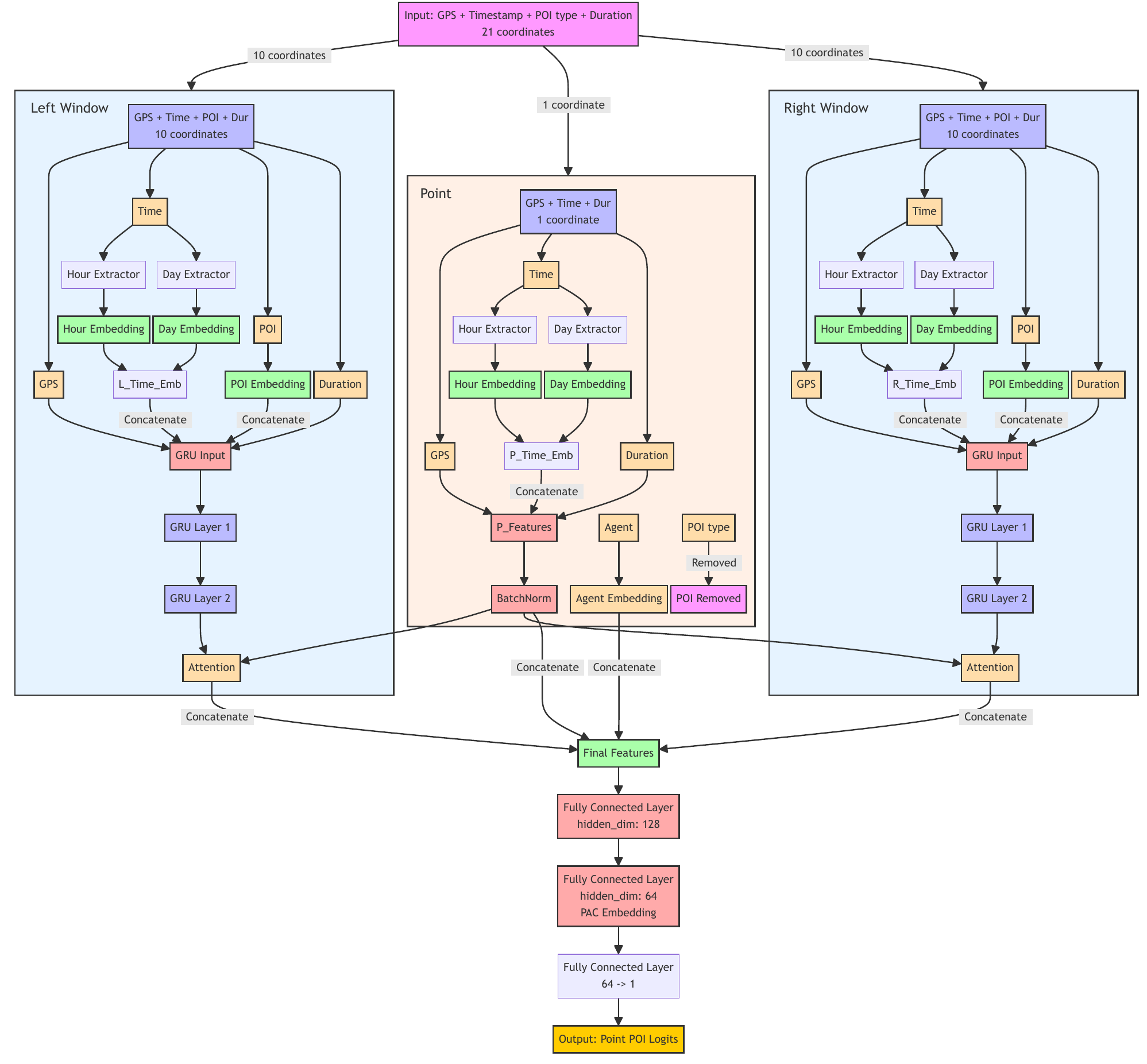}
    \caption{Illustration of the Point Activity Classifier (of STOD) adapted to the multi-agent GPS staypoint trajectory. The input to the network is 21 GPS coordinates with some extra features, which correspond to the timestamp, latitude, longitude, POI types, and stay duration of each GPS coordinate. The input is split into the left window, the coordinate of interest, and the right window. All other inputs are used to predict the POI type of the POI type of the coordinate of interest.}
    \label{fig:pac-flow}
\end{figure*}

Then, for a given GPS trajectory, it uses the concatenated PAC embeddings and Word2Vec embeddings of all points in the trajectory as input to a GRU to finally predict the bus ID. Similar to Rio-Bus, it assumes outlier bus trajectories' IDs are harder to predict. The architecture of the adapted PAC classifier is presented in Figure \ref{fig:pac-flow}.

\paragraph{Word2Vec modification}

The Word2Vec representation is compatible with agent trajectories, so we leave it unchanged. For training the Word2Vec model, we use the full trajectory of each agent as a sentence.

\paragraph{Agent ID Replacement}

In order to output the score per staypoint, we remove the GRU at the final stage and directly use a 2-layer MLP with ReLU activation to predict the agent ID, instead of the bus ID.

\paragraph{PAC modification}

We mainly modify the Point Activity Classifier, see Figure \ref{fig:pac-flow}. We remove the route ID, velocity and acceleration from the GRU input since they are not available in our dataset. Instead, we concatenate the GRU input with Points of Interest (POI) embeddings. This provides context about the surrounding environment of each GPS coordinate. Staypoint durations are also directly concatenated with the GRU input. This helps capture the temporal aspect of staypoints. Instead of predicting the bus state, we predict the POI type of the GPS coordinate of interest and make sure the POI type of the point is removed from the input.

\paragraph{Output Score}
We use the negative probability of the agent ID being correctly predicted as the anomaly score. If the agent sub-trajectory is less likely correctly identified, it is more likely to be anomalous. The agent-level anomaly score is, again, computed as the maximum of all staypoint-level anomaly score.

\subsection{GM-VSAE}

The Gaussian Mixture Variational Sequence AutoEncoder (GM-VSAE)~\cite{Liu_Zhao_Cong_Bao_2020} captures complex sequential information in trajectories, discovers different types of normal routes via Gaussian mixture, and represents them in a continuous latent space. This method requires gridified sequences. Therefore, we only retain the longitude and latitude from the input GPS sequences. We then convert these coordinates into grid series for use with the model. We treat each agent as a separate trajectory. 

This model does not support staypoint-level scoring. We calculate the negative likelihood of each test agent trajectory and use this value as the agent-level anomaly scores. 

\section{Visit Rate Model}
\label{appendix:visit_rate_model}

The visit rate model assumes that the number of times an agent $i$ visits a POI $p$ in a 4-week time window is a Poisson process with a rate parameter $\lambda_{i,p}$. We further assume that this parameter does not change between \texttt{train} and \texttt{test} for non-anomalous agents and stay points.

We then compute staypoint-level anomaly scores $a_{i,p}$ for each (agent\_id, poi\_id) tuple simply by comparing the observations $\hat{\lambda}_{i,p}$ between between \texttt{train} and \texttt{test}, normalized by the standard deviation in \texttt{train}.

\begin{equation}
    a_{i,p} = \frac{\mathrm{abs}(\hat{\lambda}_{i,p,\mathrm{train}}-\hat{\lambda}_{i,p,\mathrm{test}})}{\sqrt{\hat{\lambda}_{i,p,\mathrm{train}}}}
\end{equation}

If no visits occurred in \texttt{train}, a baseline value of 0.5 was assumed. This results in a single visit in \texttt{test} receiving a slightly anomalous score rather than causing division by zero.

Finally, since this method only picks up inserted anomalies, which are typically preceded and followed by altered anomalies, we re-assigned each staypoint's anomaly score to be the max of itself, the preceding, and the following score.

Agent-level scores were then computed by taking $\max_{p}(a_{i,p})$ for each agent.

\end{appendices}

\end{document}